\documentclass[a4paper]{article}
\usepackage{INTERSPEECH2021}
\usepackage{amsmath,graphicx}
\usepackage{arydshln}

\usepackage{float}
\usepackage{amssymb}
\usepackage{hyperref}
\usepackage{multicol}
\usepackage[flushleft]{threeparttable}

\usepackage{comment}
\usepackage[table]{xcolor}
\usepackage{xurl}
\usepackage{hyperref}
\hypersetup{colorlinks=true}

\definecolor{my_green}{HTML}{228B22}

\title{Dr-Vectors: Decision Residual Networks and an Improved Loss \\ for Speaker Recognition}
\name{Jason Pelecanos \qquad Quan Wang \qquad Ignacio Lopez Moreno}
{
\address{Google LLC, USA}
\email{\{\href{mailto:pelecanos@google.com}{pelecanos},\href{mailto:quanw@google.com}{quanw},\href{mailto:elnota@google.com}{elnota}\}@google.com}}

\begin{document}

\maketitle
\begin{abstract} 

Many neural network speaker recognition systems model each speaker using a fixed-dimensional embedding vector. These embeddings are generally compared using either linear or 2nd-order scoring and, until recently, do not handle utterance-specific uncertainty. In this work we propose scoring these representations in a way that can capture uncertainty, enroll/test asymmetry and additional non-linear information. This is achieved by incorporating a 2nd-stage neural network (known as a decision network) as part of an end-to-end training regimen. In particular, we propose the concept of decision residual networks which involves the use of a compact decision network to leverage cosine scores and to model the residual signal that's needed. Additionally, we present a modification to the generalized end-to-end softmax loss function to target the separation of same/different speaker scores. We observed significant performance gains for the two techniques.
\end{abstract}


\section{Introduction}
\label{sec:intro}


Speaker recognition techniques over the last decade have used fixed low-dimensional representations to represent the voice characteristics of speakers. Dehak~\cite{dehak_2011_1} introduced i-vectors which are based on a factor analysis of Gaussian mixture model statistics. More recent work employed discriminatively trained neural networks to generate intermediate statistics such as d-vectors~\cite{heigold_2016_1,wan_2018_1}, \textit{deep speaker} representations~\cite{li_2017_1}, x-vectors~\cite{snyder_2018_1}, and other variations~\cite{desplanques_2020_1, tong_2020_1}. Both i-vectors and x-vectors are generally followed by some linear discriminant analysis (such as LDA or PLDA~\cite{prince_2007_1}) and d-vectors utilize cosine similarity scoring. For the widely used methods, some of the current assumptions in scoring speaker representations include: (i) enrollment and verification are treated symmetrically; (ii) utterance specific uncertainties are not captured; and (iii) scoring assumptions are based on a linear/2nd-order analysis. 


To add context for the proposed work, one paper~\cite{yaman_2012_1} investigated the use of neural network bottleneck features with one difference. The bottleneck features were trained to improve on the scores of another source in a somewhat complementary/residual way. This principle is also realized with ResNets~\cite{he_2015_1}. In particular, the output of a ResNet is the combination of its input and a multi-layer network residual.

Rather than feature bottlenecks, a network could be trained directly on concatenated enroll-verify speaker embedding combinations. The idea of using what is called a \textit{decision network} to jointly model multiple inputs was previously explored for various tasks; examples include stereo image analysis~\cite{zbontar_2015_1} and image patch comparisons~\cite{zagoruyko_2015_1}.

These two techniques (complementary/residual analysis and decision networks) have the potential to address the 3 challenges within the scope of the neural network domain. These challenges were investigated in other works to varying degrees. Challenge (ii), which relates to the modeling of uncertainty information, was previously addressed for non-neural network models (for example, i-vectors and supervectors). Specifically,~\cite{borgstrom_2013_1,kenny_2013_1,cumani_2013_1,cumani_2014_1,saeidi_2013_1} examined how uncertainty information of recording specific i-vectors and supervectors could be modelled to improve performance. It was observed that improvements were more pronounced for short recordings. More recent work in~\cite{brummer_2018_1} utilized heavy-tailed PLDA and examined uncertainty modeling for i-vectors with possible extensions for neural network embeddings. Interestingly, the heavy-tailed PLDA model also relates partially to challenge (iii); it is an extension to the 2nd-order PLDA model. It was introduced to the speaker recognition community in~\cite{kenny_2010_1} and a faster version was later applied for i-vectors and x-vectors~\cite{silnova_2018_1}. In related work, utterance specific uncertainty was explicitly modeled for an x-vector speaker diarization system within a PLDA framework~\cite{silnova_2020_1}. These works model uncertainty and higher order information explicitly with various assumptions. In contrast, we propose to allow the neural network model to implicitly determine how best to represent uncertainty and other information.


In this work there are two novel aspects. The main contribution, we term \emph{decision residual networks}, builds upon two key concepts from past works: residual model analysis and decision networks. Specifically, we propose a system that optimizes for both a primary signal (for example, cosine scores derived from comparing speaker embeddings) and a decision network that captures residual information. The second contribution is a modification to the loss function proposed in~\cite{wan_2018_1} which expands the type of different-speaker trials and increases the difficulty of the training problem. Other works made the learning problem more challenging by adding a margin parameter~\cite{xiang_2019_1, deng_2019_1} or by selecting difficult trials as part of a triplet loss~\cite{schroff_2015_1, bredin_2017_1, hoffer_2015_1}.


The rest of the paper is organized as follows. Section~\ref{sec:dr_vectors} introduces Decision Residual Networks and the resulting speaker representations termed \textit{Dr-Vectors}. Section~\ref{sec:x-ge2e} provides details on the improved loss function and Section~\ref{sec:system_description} describes the overall system configuration details. The results are shown in Section~\ref{sec:results} which is followed by conclusions.

\section{Decision residual networks}
\label{sec:dr_vectors}

\begin{figure}[htb]
  \centerline{\includegraphics[width=7.5cm]{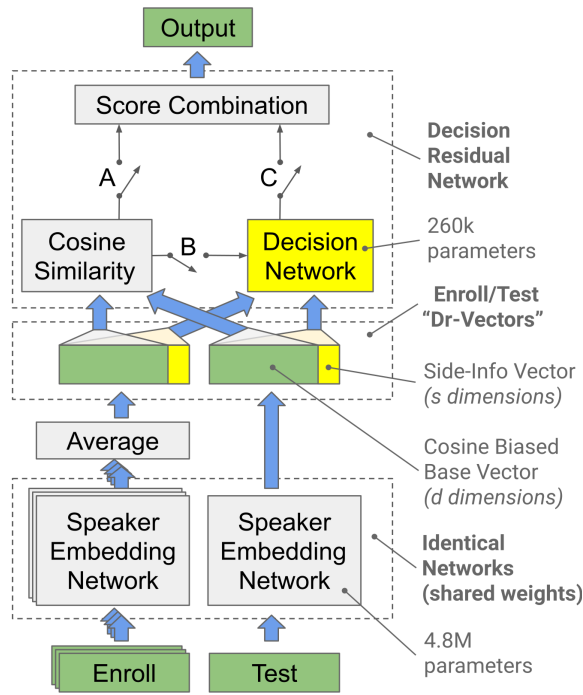}}
  \caption{Block diagram showing the speaker recognition system which is composed of speaker embedding networks, Dr-Vectors and the decision residual network. Switches $A$, $B$, and $C$ enable the evaluation of different experimental configurations through ablation studies. The parameters $d$ and $s$ configure the size of the vectors used in the cosine ($d$ dimensions) and decision network ($d+s$ dimensions) calculations. The side-info vector ($s$ dimensions) is the part of the larger speaker embedding that is not directly influenced by the cosine similarity path.}
  \label{fig:dr_vectors}
\end{figure}

To provide context, we briefly introduce components of the speaker recognition system and share details of the decision residual network. Figure~\ref{fig:dr_vectors} shows the overall system and it is composed of the \textit{speaker embedding network} structure followed by the proposed \textit{decision residual network} module. For clarity of terminology, we refer to \textit{decision residual network} to mean the combination of cosine or other similarity related information with a \textit{decision network}. In contrast, when we refer to \textit{decision network} we mean the core neural network component. The term \textit{decision network} was used in~\cite{zagoruyko_2015_1} to refer to the neural network used to produce the final similarity score and we follow that terminology here.

We first discuss the input to the decision residual network beginning at the enrollment and test recordings. The recordings are converted to a sequence of features and this sequence is transformed into a fixed-dimensional speaker embedding using a neural network. Specifics of the speaker embedding network are provided in Section~\ref{sec:system_description}. For enrollment there may be several utterances. Here, each enrollment utterance is separately passed through the speaker embedding network to produce multiple speaker embeddings. These embeddings are averaged to create the final speaker model. For the test utterance, the output of the speaker embedding network is used directly. The enrollment and test speaker embeddings, which we refer to as \textit{Dr-Vectors}, are then processed jointly by the decision residual network.

The decision residual network must be structured in a way that utilizes a decision network to capture additional residual speaker information. The challenge is, if a basic decision network is used in isolation, it would need to partially relearn the mapping that is already captured by the cosine score. The results in~\cite{yaman_2012_1} indicate that a bottleneck neural network can provide complementary features to an existing system. Also, for ResNets~\cite{he_2015_1}, the output is calculated as the combination of the output of a multi-layer neural network and its input. In this case the network estimates the needed residual to its input features. In a similar manner, we can leverage the ideas of complementary features and residuals by having a neural network estimate the residual signal to the cosine scores; \textit{i.e.}, combine information from the cosine similarity and the decision network.

The proposed system can be setup to allow for different structures within the decision residual network module. These configurations are changed by modifying $A$, $B$, $C$, and $d$. Symbols $A$, $B$, and $C$ are information switches that can be turned ``ON" (connected) or ``OFF" (disconnected). The variable $d$ specifies the number of elements in the speaker embedding vector that are used in the cosine score calculation. As an example, to represent a \textit{d-vector} type system~\cite{heigold_2016_1,wan_2018_1}, the decision network is not used (switches $A$=ON, $B$=OFF, $C$=OFF). To represent a decision residual network, the cosine score and the decision network are combined (switch $C$ is ON and at least one of the switches $A$ or $B$ are ON). The resulting speaker embeddings are termed \textit{decision residual vectors} or \textit{Dr-Vectors}.

There is also the score combination block in Figure~\ref{fig:dr_vectors}. It consists of adding the cosine score to the decision network output. This is followed by an affine transform. The affine transform (\textit{i.e.} scale and offset parameters which are also optimized as part of end-to-end training) allows the system to appropriately condition the scores for the objective function.

In closing this section, we speculate that decision residual networks are able to: (1) capture information related to enrollment and test asymmetry, (2) model recording/trial specific embedding uncertainty, and (3) handle additional non-linearity after the speaker embedding stage in the model.

\section{Generalized end-to-end extended-set softmax loss}
\label{sec:x-ge2e}

In the work presented in~\cite{wan_2018_1}, multiple Generalized End-to-End (GE2E) losses were proposed. It was observed in the paper that the GE2E softmax loss performed well for text-independent speaker recognition and we focus on it here. GE2E softmax loss performs a comparison of a sampling of test segment and enrollment embeddings as part of a mini-batch. For each test segment, we compare its embedding against a sampling of enrollment embeddings, with one enrollment embedding being from the same speaker as the test segment. This is suitable for selecting one class from a set of classes. The training criterion is similar to regular softmax cross-entropy except that it is restricted to only the randomly selected speakers chosen for each mini-batch.

In this work, we evaluate systems using the EER metric. This error metric is minimized when all same-speaker trial scores are larger than all different-speaker trial scores. With this in consideration, we apply an extension to GE2E softmax loss called \emph{GE2E eXtended-set Softmax} loss (or \emph{GE2E-XS}). Essentially, we apply a modified softmax cross-entropy calculation such that we sample from \emph{all} different-speaker trials rather than different-speaker trials that involve \emph{only} that test segment. We now suggest one possible implementation for optimization frameworks that use a fixed size/structure graph.

\begin{figure}[htb]
  \centerline{\includegraphics[width=7.5cm]{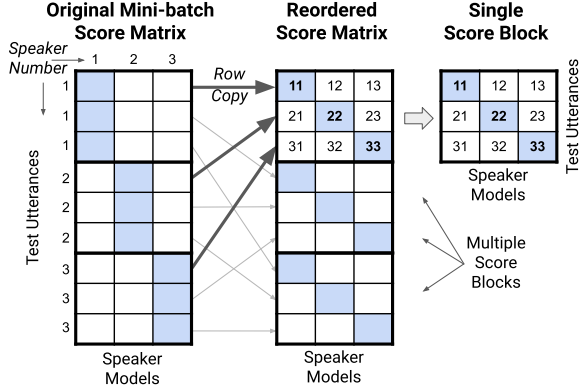}}
  \caption{Figure showing how to reorder the standard mini-batch score matrix representation into a matrix formed from the concatenation of single score blocks with same-speaker scores along the diagonal. A single score block has the diagonal elements (identified by the 3 blue shaded cells) containing same-speaker scores and the off-diagonal elements (6 unshaded cells) making up the different-speaker scores.}
  \label{fig:score_block}
\end{figure}

For training our system we establish a \textit{mini-batch score matrix} as shown on the left side of Figure~\ref{fig:score_block}. It is also implemented to be the same setup presented in~\cite{wan_2018_1}. In the figure, the shaded squares represent target scores while unshaded squares are non-target scores. The rows are then reordered to construct the \textit{reordered score matrix}. This matrix may be seen as a concatenation of multiple \textit{single score blocks}. A single ``block" consists of target scores along the diagonal elements and non-target scores in the off-diagonal elements. This particular example structure scores 3 test utterances (represented by 3 rows) against 3 speaker models (shown by the 3 columns).

The loss calculated for the mini-batch is simply the sum of the losses accumulated over the multiple \textit{single score blocks} making up the mini-batch. To simplify the notation in the equations we present the loss as a function of a \textit{single score block}. That said, let the score on the $i$th row and $j$th column be identified as $y_{ij}$ within an $N \times N$ element score block.

The regular GE2E softmax loss formulation for a single score block may be written as follows:
\begin{eqnarray}
L_{S} & = & -\sum\limits_{i} \log \frac{\exp y_{ii}}{\sum\limits_{j} \exp y_{ij}} \\
& & \mbox{where } 1 \leqslant i, j \leqslant N \nonumber
\end{eqnarray}

In contrast, the extended-set loss includes a softmax calculation based on all non-target scores within the scoring block and not just a single row. Equation~\ref{eqn:x-ge2e} shows the \textit{GE2E eXtended-set Softmax (XS)} loss for a single scoring block. As mentioned before, a mini-batch of scores can be constructed by stacking the individual score blocks. In our experiments (Section~\ref{sec:results}), each mini-batch consists of 8 stacked blocks comparing 16 test utterances against 16 speaker models.
\vspace{-0.05cm}
\begin{eqnarray}
L_{MS} & = & -\sum \limits_{i} \log \frac{\exp y_{ii}}{\exp y_{ii} + \sum \limits_{j} \sum \limits_{k \neq j} \exp y_{kj}} \label{eqn:x-ge2e} \\
& & \mbox{where } 1 \leqslant i,j,k \leqslant N \nonumber
\end{eqnarray}

\section{System configuration details}
\label{sec:system_description}

This section describes the speaker recognition system shown in Figure~\ref{fig:dr_vectors}. The work is similar to the system described in~\cite{wan_2018_1}. First we discuss the input to the speaker embedding network.

For the inputs, there are multiple enrollment recordings and a single test recording. For a single recording, the audio is partitioned into 25ms frames with a 10ms frame shift. For each frame there are 40 Mel-spaced log-filterbank-energy features covering a band of 125-3800Hz.

Next, the speaker embedding neural network consists of 3 stacked layers of LSTMs~\cite{hochreiter_1997_1} with 768 hidden nodes. The output of each LSTM layer has a dimension-reducing linear projection layer followed by a tanh activation layer~\cite{sak_2014_1}. The output of the activation layer (with 256 output nodes) is fed to the input of the next layer. The final frame of the output of the last tanh layer has a linear transformation applied and becomes the speaker embedding used as input to the decision residual network.

The decision residual network accepts as input the enrollment and test speaker representations (Dr-Vectors). The Dr-Vectors have a dimension of 256 (the same as the projected LSTM output) and are comprised of $d$ base vector dimensions (for example, 200) and $s$ side information dimensions (for example, 56). First, the cosine score is calculated from the $d$ base vector dimensions of the enrollment and test representations. Second, the input to the decision network is prepared. This involves the feature concatenation of the enrollment and test utterance representations as well as the cosine score calculated earlier (if switch ``B" is activated). This representation is fed into the decision network. The decision network is comprised of 3 linear layers with leaky-ReLU~\cite{maas_2013_1} activation functions. Each of these layers has 256 nodes and the leaky-ReLU activation is configured to scale negative input values by 0.2. The 3 layers are followed by a weighted summation operation to give a single intermediate score. 

There are two steps as part of the score combination block. Depending on which switches are activated, either one of or both the cosine score and the output of the decision network are added together. This score is scaled and an offset is applied and is optimized as part of the end-to-end training loss.  

To accomplish the end-to-end training, mini-batches are created from randomly sampled speakers. The first step is to randomly sample $16$ speakers with $8$ utterances per speaker from the entire speaker pool. This gives a total of $128$ utterances for a mini-batch. The features from each utterance are passed through the speaker embedding network to produce $128$ individual Dr-Vectors. The first $4$ vectors from each speaker are averaged to represent the speaker enrollment models while the remaining $4$ vectors are used individually as test vectors. This gives $16$ models (as before) and $4\times16=64$ test utterances. Each of these test utterances is scored against each speaker enrollment model to produce a matrix with dimensions $64\times16$. These scores are arranged into 4 score blocks of dimension $16\times16$ (see Section~\ref{sec:x-ge2e}). The diagonal in each score block contains the target trial scores, while the off-diagonal elements contain the non-target scores. These score blocks are appended to produce the $64\times16$ score matrix. The process of generating this score matrix is repeated by switching the role of enrollment and test utterances. The two matrices of dimensions $64\times16$ are appended together to give a final score matrix of $128\times16$. Each $16\times16$ block of scores within this matrix are scored according to the GE2E extended-set softmax formulation from Equation~\ref{eqn:x-ge2e}.

\section{Experimental results}
\label{sec:results}

In this section, we describe the experiment data and the corresponding results.

\subsection{Training and evaluation data}

\begin{table}[htb]
\setlength{\tabcolsep}{4.8pt}
\centering
 \begin{threeparttable}
  \caption{Composition of the training and evaluation data. All numbers are indicated in thousands (as indicated by [k]).}
  \vspace{-0.2cm}
  \label{tab:trainevaldata}
\begin{tabular}{l | c c | c c c c}
                    & \multicolumn{2}{c|}{\textbf{Training}} & \multicolumn{4}{c}{\textbf{Evaluation}} \\
        & \textbf{Spk}  & \textbf{Utt} & \textbf{Spk}  & \textbf{Utt} & \textbf{Tar} & \textbf{Non} \\
     \textbf{Language}  & [k]  & [k] & [k]  & [k] & [k] & [k] \\ \hline
  English (US)               & 4.2  & 63   & 1.4 & 203 & 189 & 200 \\ 
  Hindi                      & 5.1  & 36   & 0.7 & 26  &  21 & 200 \\ 
  Japanese                   & 4.0  & 46   & 0.8 & 54  &  49 & 200 \\ 
  Mandarin (Simp)            & 1.1  & 120  & 0.3 & 38  &  36 & 200 \\ 
  \emph{Other Data*} & 42.1 & 1404 & -    &  -    &   -   &   -   \\
\end{tabular}
\begin{tablenotes}
            \item[*] Includes data from the following languages: Cantonese, Danish, Dutch, French, German, Indonesian, Italian, Korean, Mandarin (Traditional), Norwegian, Portuguese (Brazil and Portugal), Russian, Swedish, Thai and Vietnamese. It also includes several varieties of Spanish and non-US English. 
\end{tablenotes}
\end{threeparttable}
\vspace{-0.5cm}
\end{table}

Our training set consists of vendor collected speech queries from different language varieties (see Table~\ref{tab:trainevaldata}) using devices such as laptops and cell phones. We also apply data augmentation techniques~\cite{lippmann1987multi,ko2017study,kim2017generation} involving noise sources combined with room simulation effects. Others have also applied such data augmentation techniques to i-vectors~\cite{garcia-romero_2012_1,lei_2012_1,avila_2014_1} and more recently deep neural networks~\cite{snyder_2018_1,huang_2019_1}. 

We assess evaluation performance of the systems by averaging the EER across the 4 evaluation languages shown in Table~\ref{tab:trainevaldata}. There is a clean set and a corresponding noisy set for each language. The noisy set is generated by applying noise and room simulation effects to the clean data. The noise sources and room parameters chosen for creating the noisy speech during evaluation are different from the training phase.

\subsection{GE2E extended-set softmax loss}

In this section we compare the equal-class-weight (ECW) sigmoid binary cross-entropy loss, the GE2E softmax loss and the GE2E extended-set softmax loss (Table~\ref{tab:x-ge2e_results}). Results indicate that the extended-set softmax objective function (see Section~\ref{sec:x-ge2e}) provides benefit on this task set and is used for all following results\footnote{We did not compare ECW binary cross-entropy loss and GE2E softmax loss under different configurations such as larger batch sizes, so the smaller differences between these two losses may not generalize.}. 

\begin{table}[htb]

\centering
\caption{Comparison of equal-class-weight sigmoid binary cross-entropy, GE2E softmax and GE2E extended-set softmax losses. These results are presented for cosine similarity scoring. (Metric: Average EER in \%)}
  \label{tab:x-ge2e_results}
\begin{tabular}{l | c c}
  \textbf{Loss Type} & \textbf{Clean} & \textbf{Noisy} \\ \hline
  ECW Binary Cross-Entropy & 0.95 & 2.29 \\ 
  GE2E Softmax                  & 0.96 & 2.50 \\ 
  GE2E Extended-Set Softmax        & \textbf{0.77} & \textbf{1.71} \\ 
\end{tabular}
\vspace{-0.3cm}
\end{table}

\subsection{Utility of the cosine similarity}

In this section we examine how the cosine similarity can be leveraged in concert with a decision network. To this end, we refer to Figure~\ref{fig:dr_vectors} and the 3 switches: $A$, $B$ and $C$. The configured state of these switches (``ON" or ``OFF") determines how the cosine scores are used and if the decision network is trained. In Table~\ref{tab:switchedresults} we share the results for different on/off configurations. For example, the first row of results gives the error rates for the cosine score only system (which is also the result from Table~\ref{tab:x-ge2e_results}). The second row shows the results for the decision network by itself. The third row of results represents the system which uses the decision network but also includes the cosine score as an input feature to the network. The next row adds scores from the decision network (without the cosine score as input) and the cosine score directly. The final row enables all connections.

An important observation is that using both the cosine score and the decision neural network is better than using either component alone. By itself, the decision network is comparable in performance to cosine scoring. Additionally, we note that adding the residual network increased the total parameters by less than 6\%. In contrast, we found that increasing the number of parameters in the LSTM layers by 6\% had a negligible impact on performance.

\begin{table} [htb]
\centering
  \caption{This table shows the results for different configurations of the system shown in Figure~\ref{fig:dr_vectors}. Specifically, it explores different ways the cosine score can be utilized with the decision network. Here $d$ is set to 200 throughout except for the first row of results where it makes sense to use all terms ($d=256$) for cosine scoring. (Metric: Average EER in \%)}
  \label{tab:switchedresults}
\begin{tabular}{c c c | c c}
  \multicolumn{3}{c|}{\textbf{Switch States}} & & \\ 
  $\textbf{\textit{A}}$ & $\textbf{\textit{B}}$ & $\textbf{\textit{C}}$ & \textbf{Clean} & \textbf{Noisy} \\ \hline
  ON  & OFF & OFF & 0.77 & 1.71 \\ 
  OFF & OFF & ON  & 0.74 & 1.70 \\ 
  OFF & ON  & ON  & 0.67 & 1.48 \\ 
  ON  & OFF & ON  & 0.68 & 1.45 \\ 
  ON  & ON  & ON  & \textbf{0.66} & \textbf{1.43} \\ 
\end{tabular}
\vspace{-0.3cm}
\end{table}

\subsection{Number of terms in the cosine similarity calculation}

We also study variations in performance with the number of Dr-Vector embedding parameters used in the cosine score calculation. To that end, we perform experiments by changing the number of elements $d$ (see Figure~\ref{fig:dr_vectors}). The results are included in Table~\ref{tab:numtermsresults}. For this experiment the best overall results were observed for $d=200$.

\begin{table} [htb]
\centering
  \caption{Examining Dr-Vector performance as a function of, $d$, the number of terms used in the cosine calculation. Here we set ($A$=ON, $B$=ON, $C$=ON). (Metric: Average EER in \%)}
  \label{tab:numtermsresults}
\begin{tabular}{l | c c}
  \textbf{\# Terms, }$\textbf{\textit{d}}$ & \textbf{Clean} & \textbf{Noisy} \\ \hline
  \,\,\,\, 0 \, (none) & 0.74 & 1.70 \\ 
  128 (partial)        & 0.70 & 1.55 \\ 
  200 (partial)        & \textbf{0.66} & \textbf{1.43} \\ 
  256 \,\,\, (all)     & 0.68 & 1.49 \\ 
\end{tabular}
\vspace{-0.3cm}
\end{table}

\section{Conclusions}

In this work we showed that decision residual networks can significantly improve speaker recognition system performance while only marginally increasing the number of model parameters required. The proposed model can generate speaker embeddings called Dr-Vectors which can implicitly carry additional information pertinent for scoring. It is speculated that such information could include enroll/test asymmetry statistics, uncertainty, and non-linearities. Additionally, we proposed an extension to the GE2E softmax loss which more broadly samples from the non-target distribution to enhance performance. We note that similar ideas can be applied to other speaker recognition deep learning frameworks.

\section{Acknowledgements}

The authors are appreciative of David Nahamoo for providing awareness in 2012 of what was later termed \textit{decision networks}. We thank Niko Br\"{u}mmer and the reviewers for their feedback.

\clearpage
\vfill\pagebreak

\bibliographystyle{IEEEtran}
\bibliography{refs}

\end{document}